\documentclass{article}

\usepackage[nonatbib,final]{nips_2017}

\usepackage[utf8]{inputenc} 
\usepackage[T1]{fontenc}    
\usepackage{hyperref}       
\usepackage{url}            
\usepackage{booktabs}       
\usepackage{amsfonts}       
\usepackage{nicefrac}       
\usepackage{microtype}      
\usepackage{url}
\usepackage{latexsym}
\usepackage{lipsum}
\usepackage{mathtools}
\usepackage{amssymb}
\usepackage{hyperref}
\usepackage{array}
\usepackage{float}

\title{Multi-range Reasoning for Machine Comprehension}

\author{
  Yi Tay$^1$, Luu Anh Tuan$^2$, and Siu Cheung Hui$^3$\\
  $^{1,3}$Nanyang Technological University\\
  $^2$Institute for Infocomm Research \\
  \texttt{ytay017@e.ntu.edu.sg$^1$}\\
  \texttt{at.luu@i2r.a-star.edu.sg$^2$}\\
  \texttt{asschui@ntu.edu.sg$^3$} \\
}

\begin{document}

\maketitle

\begin{abstract}
 We propose \textsc{MRU} (Multi-Range Reasoning Units), a new fast compositional encoder for machine comprehension (MC).
Our proposed \textsc{MRU} encoders are characterized by multi-ranged gating, executing a series of parameterized contract-and-expand layers for learning gating vectors that benefit from long and short-term dependencies. The aims of our approach are as follows: (1) learning representations that are concurrently aware of long and short-term context, (2) modeling relationships between intra-document blocks and (3) fast and efficient sequence encoding. We show that our proposed encoder demonstrates promising results both as a standalone encoder and as well as a complementary building block. We conduct extensive experiments on three challenging MC datasets, namely RACE, SearchQA and NarrativeQA, achieving highly competitive performance on all. On the RACE benchmark, our model outperforms DFN (Dynamic Fusion Networks) by $1.5\%-6\%$ without using any recurrent or convolution layers. Similarly, we achieve competitive performance relative to AMANDA \cite{kundu2018amanda} on the SearchQA benchmark and BiDAF \cite{seo2016bidirectional} on the NarrativeQA benchmark without using any LSTM/GRU layers. Finally, incorporating MRU encoders with standard BiLSTM architectures further improves performance, achieving state-of-the-art results. 
\end{abstract}

\section{Introduction}
Teaching machines to read, comprehend and reason lives at the heart of machine comprehension (MC) tasks \cite{rajpurkar2016squad,lai2017race,dunn2017searchqa,kovcisky2017narrativeqa}. In these tasks, the goal is to answer questions based
on a given passage, effectively testing the learner's capability to understand natural language. This has been an extremely productive area of research in the recent years, giving rise to many highly advanced neural network architectures \cite{DBLP:journals/corr/XiongZS16,seo2016bidirectional,hu2017mnemonic,shen2017reasonet,wang2016machine}. A common denominator in many of these models is the compositional encoder, i.e., usually a bidirectional recurrent-based (LSTM \cite{hochreiter1997long} or GRU \cite{cho2014learning}) encoder that sequentially parses the text sequence word-by-word. This helps to model compositionality of words, capturing rich and complex linguistic and syntactic structure in language. 

While the usage of recurrent encoder is often regarded as indispensable in highly complex MC tasks, there are still several challenges and problems pertaining to it's usage in modern MC tasks. Firstly, documents can be extremely long to the point where running a BiRNN model across a long document is computationally prohibitive. This is aggravated since MC tasks can be easily extended to reasoning over multiple long documents. Secondly, recurrent encoders have limited access to long term context since each word is sequentially parsed. This restricts any form of multi-sentence and intra-document reasoning from happening within compositional encoder layer. 

To this end, we propose a new compositional encoder that can either be used in-place of standard RNN encoders or serve as a new module that is complementary to existing neural architectures. Our proposed \textsc{MRU} encoders learns gating vectors via multiple contract-and-expand layers at multiple dilated resolutions. Specifically, we compress the input document an arbitrary $k$ times at multi-ranges (e.g., $1,2,4,10,25$) into a neural bag-of-words (summed) representation. The compact sequence is then passed through affine transformation layers and then re-expanded to the original sequence length. The $k$ document representations (at multiple ranges and n-gram blocks) are then combined and modeled with fully connected layers to form the final compositional gate which are applied onto the original input document. This can be interpreted as compositional gating by exploiting information at multiple-ranges, modeling relationships across different granularities and hierarchies. Intuitively, this is because 1-gram blocks are compared with 2-gram blocks and 10-gram blocks and so on.

This has several advantages. Firstly, we enable a major speedup by avoiding either costly step-by-step gate construction while still maintaining interactions between neighboring words. As such, our model belongs to a class of architectures which is inspired by QRNNs \cite{DBLP:journals/corr/BradburyMXS16} and SRUs \cite{lei2017training}. The key difference is that our gates are not constructed by convolution layers but explicit block-based matching across multiple ranges. Secondly, modeling at a long range (e.g., 25 or 50) enables our model to look further ahead as opposed to only one step forward. As such, the learned gastes possess not only information about nearby words but also a larger overview of the context. This is in similar spirit to self-attention, albeit executing within the encoder. Thirdly, the final gates are formed by modeling relationships between multi-range projections (n-gram blocks), allowing for fine-grained intra-document relationships to be captured. The overall contributions of our work is as follows:
\begin{itemize}

\item We propose \textsc{MRU} (\textbf{M}ulti-range \textbf{R}easoning \textbf{U}nits), a new compositional encoder which construct gates from a novel contract-and-expand operation. We propose an overall architecture that utilizes \textsc{MRU} within a bi-attentive framework for both multiple choice and span prediction MC tasks. MRU can be used as a standalone (without RNNs) for fast reading and/or together with RNN models (i.e., MRU-LSTM) for more expressive reading. 
\item We conduct extensive experiments on three large-scale and challenging machine comprehension datasets - RACE \cite{lai2017race}, SearchQA \cite{dunn2017searchqa} and NarrativeQA \cite{kovcisky2017narrativeqa}. Our model is lightweight, fast and efficient, achieving state-of-the-art or highly competitive performance on all benchmarked datasets. Since MC datasets often require a considerable amount of reasoning and natural language understanding, we believe that they serve as good testbeds for benchmarking encoders.
\item  On RACE, our model outperforms Dynamic Fusion Networks (DFN) \cite{xu2017towards}, a highly complex model. While DFN takes approximately a week to train, spending at least several hours per epoch, our model converges in less than $12$ hours with only $4-5$ minutes per epoch. Moreover, our model outperforms DFN by $2\%-6\%$ on the RACE benchmark and other strong baselines such as the Gated Attention Reader by $10\%$. On RACE, we outperform DFN without any recurrent and convolution layers. Ablation studies show an improvement of up to $6\%$ when using MRU over a LSTM/GRU encoder.

\item On the recent SearchQA benchmark \cite{dunn2017searchqa}, we achieve competitive performance relative to AMANDA \cite{kundu2018amanda}, a state-of-the-art model without using any recurrent or convolution layers. Our model runs at $2$ minutes per epoch, approximately five times faster than AMANDA. Incorporating our MRU block with standard BiLSTM architectures (MRU-LSTM) outperforms AMANDA by a reasonable margin.
\item On the NarrativeQA benchmark (summaries setting) \cite{kovcisky2017narrativeqa}, our MRU encoders achieves highly competitive performance relative to BiDAF \cite{seo2016bidirectional} a strong MC baseline without using any LSTM/GRU layers. On the other hand, MRU-LSTM significantly outperforms BiDAF, achieving state-of-the-art performance on this dataset. 
\end{itemize}

\section{Our Proposed \textsc{MRU} Encoder}

\begin{figure}[ht]
  \centering
    \includegraphics[width=0.9\linewidth]{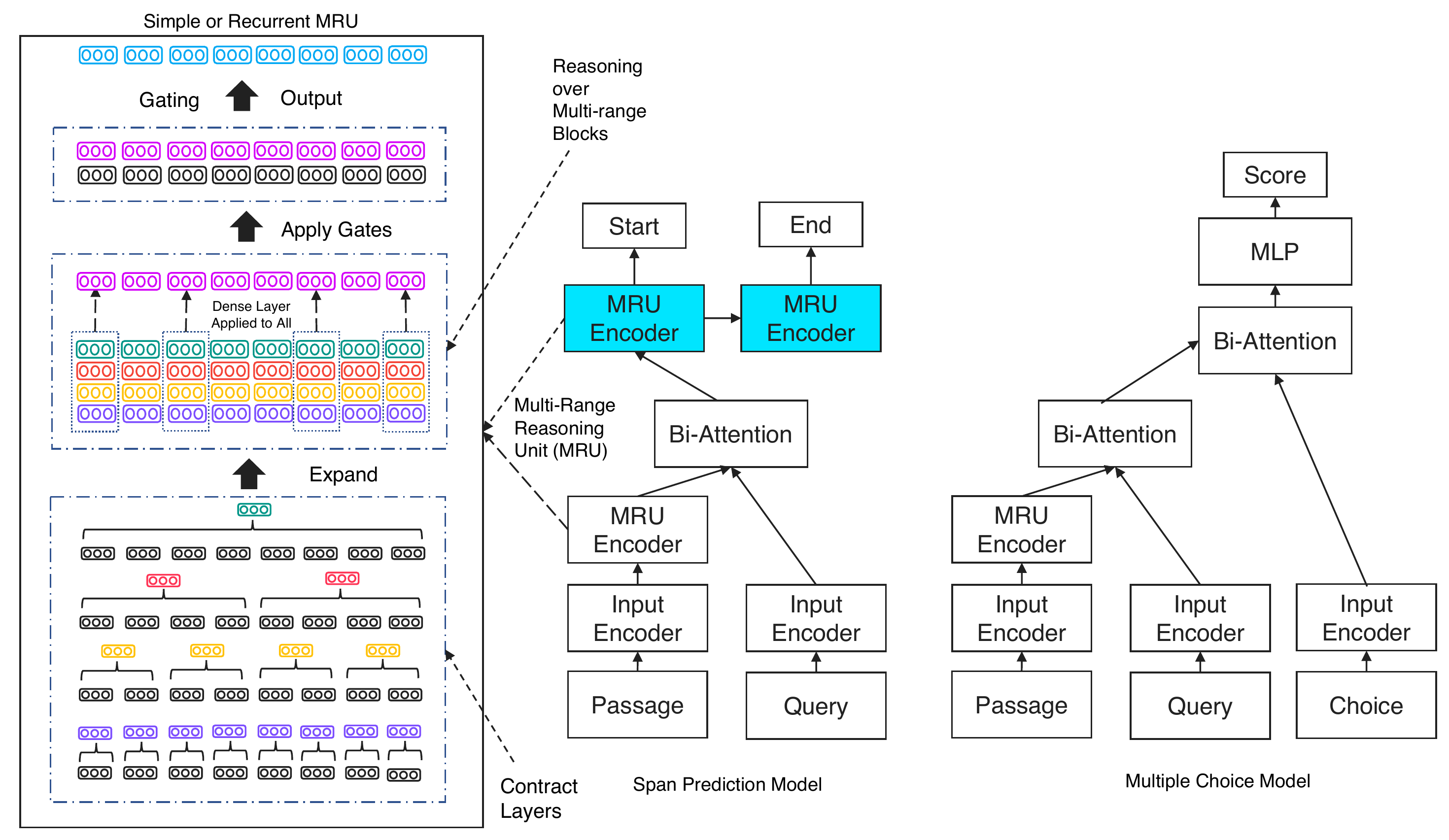}
    \caption{High-level overview of our proposed \textsc{MRU} encoder (left) and Bi-Attentive architecture for two types of MC tasks. MRU illustration shows contraction values of $\{1,2,4,8\}$. Documents are contracted and then projected with affine transformations. Subsequently, they are expanded to the original document length. A new projection layer compares the representations across multiple ranges. This multi-range representation is used as a gate to influence the input sequence.}

    \label{fig:high_level}
\end{figure}

In this section, we describe our proposed MRU encoder. The inputs to the MRU encoder is an input document $\{ w_1, w_2 \cdots w_{\ell} \}$, and list of ranges $\{r_1, r_2 \cdots r_{k}\}$ where $k$ is the number of times the contract and expand operation is executed. The final output of the encoder is a sequence of vectors which retain the same dimensionality as its inputs. Figure \ref{fig:high_level} (left most block) provides an illustration of the overall encoder architecture.

\subsection{Contract-and-Expand Operation}
This section describes the operation for each $r_{j}$. For the sake of brevity, we drop the superscripts $j$. For each $r_{j}$ and the input document, the contract operation performs takes the summation of every $r_{j}$ words. This reduces the overall document length to
$\ell / r_{j}$ where each item in the sequence is the sum of every $r_{j}$ words. Given the new sequence of $\ell / r_{i}$ tokens, we then pass each token into a single layered feed-forward neural network:
\begin{align}
\bar{w}_{t} = \sigma_{r}(\textbf{W}_a(w_{t})) + \textbf{b}_a
\end{align}
where $\textbf{W}_{a} \in \mathbb{R}^{d \times d}$ and $\textbf{b}_{a} \in \mathbb{R}^{d}$ are the parameters of the contract layer. $\sigma_r$ is the ReLU activation function. $w_t$ is the t-th token in the sequence. Given the transformed tokens $\bar{w}_1, \bar{w}_2 \cdots \bar{w}_{\ell / r_{j}}$, we then expand them into the original sequence length. Note that for each $r_{j}$, the parameters $\textbf{W}_{a}, \textbf{b}_{a}$ are not shared.

\subsection{Reasoning over Multi-ranged Blocks}
From $k$ different calls of the Contract-and-Expand operation at different ranges, we pass the concatenated vector of all transformed tokens into a two layered feed-forward neural network. 
\begin{align}
g_{t} = \textbf{F}_{2}(\textbf{F}_{1}([w^{1}_{t}; w^{2}_{t}; \cdots w^{k}_{t}]))
\label{gt}
\end{align}
where $\textbf{F}_1(.), \textbf{F}_2(.)$ are feed-forward networks with ReLU activations, i.e., $\sigma_{r}(W_{x} + b)$. $[;]$ is the concatenation operator. $g_{t}$ is interpreted as a gating vector learned from multiple ranges and Equation (\ref{gt}) is learning the relationships between a token's representation at multiple hierarchies depending on the values of $r_j$.  Notably, it is easy to see that every $n$ pairs of words will have the same gating vector where $n$ is the lowest value of $r_j$. As such, the value of the 1gram, i.e., $r_j=1$ (projection of every single token) is critical as it prevents identical gating vectors across the sequence.

\subsection{MRU Encoding Operation}
To learn the \textsc{MRU} encoded representation of each word, we consider two variations of MRU encoders. 
\subsubsection{Simple MRU}
In this variation, we use $g_t$ as a gating vector to control the fine-grained balanced between the projection of each word $w_t$ in the original input document and the original representation. 
\begin{align}
z_{t} &= tanh(\textbf{W}_{p} \: w_{t}) + \textbf{b}_{p} \\
y_{t} &= \sigma(g_{t}) * w_{t} + (1-\sigma(g_{t})) \: z_{t}
\end{align}
where $\{y_{1}, y_{2}, \cdots y_{\ell}\}$ is the output document representation. $\sigma$ is the sigmoid function. Note that this formulation is in similar spirit to highway networks \cite{DBLP:journals/corr/SrivastavaGS15}. However, since our gating function is learned via multi-range reasoning, it captures more compositionality and long range context. Note that an optional and additional projection may be applied to $w_{t}$ but we found that it did not yield much empirical benefit. 

\subsubsection{Recurrent MRU}
In the second variation, we consider a recurrent (sequential) variant. This is in similar spirit to QRNNs \cite{DBLP:journals/corr/BradburyMXS16} and SRUs \cite{lei2017training} which reduces computation cost by pre-learning the gating vectors. The following operations describe the operations of the recurrent MRU cell for each timestep $t$.
\begin{align}
c_{t} &= g_{t} \odot c_{t-1} + (1-g_{t}) \odot z_{t} \\
h_{t} &= o_{t} \odot c_{t}
\end{align}
where $c_{t}, h_{t}$ are the cell and hidden states at time step $t$. $g_{t}$ are the gates learned from out multi-range reasoning step. $o_{t}$ is an additional output gate learned via applying an affine transform on the input vector $w_{t}$, i.e., $o_{t} = W_{o}(w_{t}) + b_{o}$. Similar to RNNs, the Recurrent MRU parses the input sequence word-by-word. However, the cost is significantly reduced because we do not have expensive matrix operations that are executed in a non-parallel fashion. Finally, the outputs of the MRU encoder are a series of hidden vectors $\{h_{1}, h_{2} \cdots h_{\ell}\}$ for each word in the sequence. 

\section{Overall Model Architectures}
This section describes the overall model architecture that utilizes \textsc{MRU} encoders. In our experiments, we focus on both multiple-choice based (RACE) and span prediction MC tasks (SearchQA, NarrativeQA). Since the core focus of this paper is our encoder, we briefly provide the high-level details of our vanilla Bi-Attentive model. The Bi-Attentive models that are used in our experiments act as baselines, often being less complex than current competitive models such as BiDAF \cite{seo2016bidirectional}, AMANDA \cite{kundu2018amanda} or DFN \cite{xu2017towards}. 

\subsection{Multiple Choice Models}
In MCQ models, there are three types of input sequences, namely Passage ($P$), Question ($Q$) and Answers ($A_j$). The output of the model (for each answer), is a score $s(P,Q,A_j) \in [0,1]$ denoting the strength of $A_j$. The problem is formulated as a listwise approach, in which multiple answers are modeled concurrently with respect to $P,Q$. 

\begin{itemize}
\item \textbf{Input Encoding} - Each input sequence is passed into first a projection layer. To enhance the input word representations, we also include the standard EM (exact match) binary feature to each word. In this case, we use a three-way EM adaptation, i.e., $EM(P,Q), EM(Q,A)$  and $EM(P,A)$. The projected embeddings are then passed into a single-layered highway network. 
\item \textbf{Compositional Encoder} - In our experiments, we vary the encoder in this layer. Typical choices of encoders in this layer are LSTMs or GRUs. We vary this in our experiments in order to benchmark the effectiveness of our proposed MRU encoder. The output of this layer is same dimensions as its inputs (typically the hidden states of a RNN model). 
\item \textbf{Bi-Attention Layer} - This layer models the interactions between $P,Q$ and $A$. Let $B(.)$ be a standard bidirectional attention that utilizes mean-pooling aggregation. The scoring function is the bilinear product of the nonlinearly transformed input i.e., $F(x)^{\top}_{i}\textbf{M}F(y)_{i}$. We first apply $B(P,Q)$ to form bi-attentive $P^{q},Q^{p}$ representations. Subsequently, we apply $B(P^{q},A_j)$ to learn a vector representation for each answer. A temporal sum pooling is applied on the outputs of $P^{qa}, A_j^{p}$ and concatenated to form $a^{f}_j \in \mathbb{R}^{2d}$.  
\item \textbf{Answer Selection}
Let $\{ a_1, a_2 \cdots a_{N_{a}} \}$ be the inputs to this layer and $N_{a}$ is the number of answer candidates. Motivated by work in retrieval-based QA \cite{DBLP:conf/sigir/SeverynM15,DBLP:conf/sigir/TayPLH17}, we include word overlap features to each answer candidate. This word overlap feature is in similar spirit to the EM feature. Each overlap operation between two sequence returns four features.  We convert each answer vector $a_j$ into a scalar via $a^{f}_{j} = Softmax(\textbf{W}_{2}(\sigma_{r}(\textbf{W}_{1}([a_{j}])+b_{1})+b_{2}))$. 
\end{itemize}
The MCQ-based model minimizes the multi-class cross entropy where the number of classes corresponds to the number of choices. 
\subsection{Span Prediction Model}
Span prediction models models the relationship between $P$ and $Q$. The goal is to extract (or predict a span $s,e$) where $P[s:e]$ is the answer to the query. For most part, the model architecture remains similar especially for the input encoding layers compositional encoder layer. The key difference is that we reduce the number of input sequence from three to two. 
\begin{itemize}
\item \textbf{Input Encoding} - This follows the same design as the MCQ model, albeit for two sequence. Similarly, the two-way EM feature is added before passing into the highway layer. 
\item \textbf{Compositional Encoder} - This remains identical as the MCQ-based model.
\item \textbf{Bi-Attention Layer} - We adopt a different bi-attention function for span prediction. More specifically, we use the `SubMultNN' or the `'Mult' adaptation from \cite{DBLP:journals/corr/WangJ16b} (this is tuned) and compare aligned sequences between $P$ and $Q$ to form $P^{q}$, the query-dependent passage representation. 
\item \textbf{Answer Pointer Layer} - In this layer, we pass $P^{q}$ through a two layered compositional encoder (which is varied). The start pointer and end pointer is determined by $F(H_{1}),F(H_{2})$ where $H_{1}, H_{2}$ are the hidden outputs from the first and second encoder respectively. $F(.)$ is a linear transform, projecting each hidden state to a scalar. We pass both of them into softmax functions to obtain probability distributions. 
\end{itemize}
Following \cite{seo2016bidirectional,wang2016machine}, we minimize the joint cross entropy loss of the start and end probability distributions. During inference, finding the best answer span follows \cite{wang2016machine}.

\section{Empirical Evaluation}
In this section, we report our experimental results and comparisons against other published work. 

\subsection{Datasets}
For our experiments, we use one challenging multiple choice MC dataset and two span-prediction MC datasets.
\begin{itemize}
\item \textbf{RACE} (Reading Comprehension from Examinations) \cite{lai2017race} is a recently proposed dataset that is constructed from real world examinations. Given a passage, there are several questions with four options each. The authors argue that RACE is more challenging compared to popular benchmarks (e.g., SQuAD \cite{rajpurkar2016squad}) as more multi-sentence and compositional reasoning is required. There are two subsets of RACE, namely RACE-M (Middle school) and RACE-H (High school).  
\item \textbf{SearchQA} \cite{dunn2017searchqa} is a recent dataset that emulates a real world QA system. It involves extracting passages from search engine results and require models to answer questions by reasoning and reading these search snippets. 
\item \textbf{NarrativeQA} \cite{kovcisky2017narrativeqa} is a recent benchmark proposed for story-based reading comprehension. Different from many MC datasets, the answers are handwritten by human annotators. 
\end{itemize}

MCQ datasets are evaluated using the standard accuracy metric. For RACE, we train models on the entire dataset, i.e., both RACE-M and RACE-H and evaluate separately. For RACE, the model selection is based on each subset's respective development set. For SearchQA, we follow \cite{kundu2018amanda,dunn2017searchqa} which evaluates unigram exact match (EM) and n-gram F1 scores. For NarrativeQA, since the answers are human written and not constrained to spans in the passage, the evaluation metrics are Bleu-1, Bleu-4, Meteor and Rouge-L following \cite{kovcisky2017narrativeqa}. 

\subsection{Competitor Methods}
We describe the key competitors on each dataset. 
\begin{itemize}
\item \textbf{RACE} - the key competitors are the Stanford Attention Reader (Stanford AR) \cite{chen2016thorough}, Gated Attention Reader (GA) \cite{dhingra2016gated}, and Dynamic Fusion Networks (DFN) \cite{xu2017towards}. GA incorporates a multi-hop attention mechanism that helps to refine the answer representations. DFN is an extremely complex model. It uses BiMPM's matching functions \cite{DBLP:conf/ijcai/WangHF17} for extensive matching between $Q,P$ and $A$, multi-hop reasoning powered by ReasoNet \cite{shen2017reasonet} and employs reinforcement learning techniques for dynamic strategy selection.

\item \textbf{SearchQA} - the main competitor baseline is the AMANDA model proposed by \cite{kundu2018amanda}. AMANDA uses a multi-factor self-attention module, along with a question focused span prediction. AMANDA also uses BiLSTM layers for input encoding and at the span prediction layers. We also compare against the reported ASR \cite{kadlec2016text}  baselines which was reported in \cite{dunn2017searchqa}. 

\item \textbf{NarrativeQA} - On the NarrativeQA benchmark, we compare with the reported baselines in \cite{kovcisky2017narrativeqa}. We compete on the summaries setting, in which the baselines are a context-less sequence to sequence (seq2seq) model, ASR \cite{kadlec2016text} and BiDAF \cite{seo2016bidirectional}. 
\end{itemize}
\subsection{Our Methods}
Across our experiments, we benchmark several variants of our proposed MRU. The first is denoted as Sim. MRU which corresponds to the Simple MRU model described earlier. The model denoted by MRU (without any prefix) corresponds to the recurrent MRU model. Finally, the final variant is the MRU-LSTM which places a MRU encoder layer on top of a BiLSTM layer. We report the dimensions of the encoder as well as training time (per epoch) for each variant. The encompassing framework for MRU is the Bi-Attentive models described for MCQ-based problems and Span prediction problems. Unless stated otherwise, the encoder in the pointer layer for span prediction models also uses MRU. However, for the Hybrid MRU-LSTM models, answer pointer layers use BiLSTMs. For the RACE-dataset, we additionally report scores of an ensemble of nine Sim. MRU models. This is to facilitate comparison against ensemble models of \cite{xu2017towards}. 
\begin{table*}[t]
  \centering
 \small
    \begin{tabular}{|l|ccc|c|}
    \hline
        Model  & \multicolumn{1}{c}{RACE-M} & \multicolumn{1}{c}{RACE-H} & \multicolumn{1}{c|}{RACE} & \multicolumn{1}{c|}{Time} \\
          \hline
    Sliding Window \cite{lai2017race} & 37.3  & 30.4  & 32.2     & N/A  \\
    Stanford AR \cite{chen2016thorough} & 44.2  & 43.0    & 43.3         &  N/A\\
    GA  \cite{dhingra2016gated}  & 43.7  & 44.2  & 44.1         & N/A  \\
    ElimiNet \cite{sparikh2018eliminet} & N/A & N/A & 44.5 & N/A \\
    Dynamic Fusion Network \cite{xu2017towards}  & 51.5  & 45.7  & 47.4        & $\approx$8 hours (1 week$^\ast$) \\
    
    \hline
    BiAttention (No Encoder) & 50.6  & 44.0    & 44.9            & 3 min (9 hours)  \\
    BiAttention ($250d$ GRU) & 48.5& 42.1& 44.0 & 16 min (2 days)\\ 
    BiAttention ($250d$ LSTM) & 50.3 & 40.9 & 43.6  & 18 min (2 days)\\
    BiAttention ($250d$ Sim. MRU) & \textbf{57.7} & \underline{47.4}  & \textbf{50.4}  & 4 min (12 hours) \\
    BiAttention ($250d$ \textsc{MRU}) & \underline{56.1}  & \textbf{47.5}  &  \underline{50.0}  & 12 min (20 hours) \\
    \hline
    GA + ElimiNet \cite{sparikh2018eliminet} & N/A & N/A & 47.2  & N/A \\
    DFN Ensemble (x9) \cite{xu2017towards} & 55.6  &   49.4    & 51.2           &  N/A \\
    BiAttention (\textsc{MRU}) Ensemble (x9) & \textbf{60.2} & \textbf{50.3}       & \textbf{53.3}         &  N/A \\
    \hline
    \end{tabular}%
     \caption{Comparison against other published models on RACE dataset \cite{lai2017race}. Competitor result are reported from \cite{lai2017race,xu2017towards}. Best result for each category (single and ensemble) is in boldface. Last column reports estimated training time per epoch and total time for convergence. $^\ast$ estimated values that we obtain from asking the authors. }
  \label{tab:race}%
\end{table*}%

\subsection{Implementation Details}
We implement all models in TensorFlow \cite{tensorflow2015-whitepaper}. Word embeddings are initialized with $300d$ GloVe \cite{DBLP:conf/emnlp/PenningtonSM14} vectors and are not fine-tuned during training. Dropout rate is tuned amongst $\{0.1, 0.2,0.3\}$ on all layers including the embedding layer. For our MRU model, we use a range values of $\{1,2,4,10,25\}$. MRU encoders are only applied on the passage and not the query. We adopt the Adam optimizer \cite{DBLP:journals/corr/KingmaB14} with a learning rate of $0.0003/0.001/0.001$ for RACE/SearchQA/NarrativeQA respectively. The batch size is set to $64/256/32$ accordingly. The maximum sequence lengths are $500/200/1100$ respectively. For NarrativeQA, we use the Rouge-L score to find the best approximate answer relative to the human written answer for training the span model. All models are trained and all runtime benchmarks are based on a TitanXP GPU.

\subsection{Experimental Results on RACE}
Table \ref{tab:race} reports our results on the RACE benchmark dataset. Our proposed MRU model achieves the best result for both single models and ensemble models. We outperform highly complex models such as DFN. We also pull ahead of other recent baselines such as ElimiNet and GA by at least $5\%$. The best single model score from RACE-H and RACE-M alternates between Sim. MRU and MRU. Overall, there is a $6\%$ improvement on the RACE-H dataset and $1.8\%$ improvement on the RACE-M dataset. Our Sim. MRU model also runs at 4 min per iteration, which is dramatically faster and simpler than DFN or other recurrent models. We believe that this finding highlights the importance of designing strong and fast baselines for the task at hand. 

In general, we also found that the usage of a recurrent cell is not really crucial on this dataset since (1) Sim. MRU and MRU can achieve comparable performance to each other, (2) GRU and LSTM models do not have a competitive edge and (3) Using no encoder already achieves comparable\footnote{Nevertheless, this suggests the importance of benchmarking good and strong baselines since a well-tuned baseline model can outperform DFN, a highly complicated model. } performance to DFN. Finally, an ensemble of Sim. MRU models achieve state-of-the-art performance on the RACE dataset, achieving and overall score of $53.3\%$.

\begin{table*}[t]
  \centering
\small
    \begin{tabular}{|l|cccc|c|}
    \hline
          & \multicolumn{2}{c}{Dev} & \multicolumn{2}{c|}{Test} &        \\
         \cline{1-6}
         Model & \multicolumn{1}{c}{Acc} & \multicolumn{1}{c}{F1} & \multicolumn{1}{c}{Acc} & \multicolumn{1}{c|}{F1}  & Time  \\
          \cline{1-6}
    TF-IDF max \cite{dunn2017searchqa} & 13.0   & N/A & 12.7  &  N/A & N/A\\
    ASR \cite{kadlec2016text}  & 43.9  & 24.2  & 41.3  & 22.8  & N/A \\
    AMANDA \cite{kundu2018amanda} & 48.6  & 57.7  & 46.8  & 56.6 &  $\approx$8$^{\ast}$ min \\
    \hline
    Bi-Attention$^{\dagger}$ (No Encoder) & 12.4& 20.2 & 18.9& 12.3 &$\approx$17 sec\\

        Bi-Attention$^{\dagger}$  ($150d$ BiLSTM) & 40.0& 51.3 & 38.6& 49.0 & $\approx$7 min \\
    Bi-Attention$^{\dagger}$  ($300d$ LSTM) & 40.3& 48.7& 38.2& 46.4  & $\approx$6 min \\
 
     \hline
Bi-Attention$^{\dagger}$  ($300d$ Sim. MRU) &   44.1  &  45.5 &   42.9  &  43.1 & $\approx$25 sec\\
    Bi-Attention$^{\dagger}$  ($300d$ MRU) & 48.6 & 54.8  & 46.8 & 53.3  & $\approx$2 min \\
    \hline
    Bi-Attention ($200d$ Hybrid MRU-LSTM) & \textbf{50.5} & \textbf{59.9}   & \textbf{49.4} & \textbf{59.5}  & $\approx$7 min \\
    \hline
    \end{tabular}%
  
    \caption{Experimental Results on SearchQA dataset. \cite{dunn2017searchqa}. Unigram Accuracy and N-gram F1 are reported following \cite{kundu2018amanda}. All models with $^{\dagger}$ use the same encoder in the answer pointer layer. $^\ast$ are estimates running a replicated model with same batch size ($b=256$) as our models.}
    \label{tab:searchqa}%
\end{table*}%

\subsection{Experimental Results on SearchQA}
Table \ref{tab:searchqa} reports our results on the SearchQA dataset. We draw the reader's attention to the performance of the $300d$ MRU encoder. We achieve the same accuracy as AMANDA without using any LSTM or GRU encoder. This model runs at $2$ min per epoch, making it $4$ times more efficient than AMANDA (estimated, with identical batch size). While, AMANDA also uses multi-factor self-attention, along with character enhanced representations, our simple MRU encoder used within a mere baseline bi-attentive framework comes close in performance. Finally, the hybrid combination, MRU-LSTM significantly outperforms AMANDA by $3\%$.

Contrary to MCQ-based datasets, we found that Sim. MRU model could not achieve comparable results to the recurrent MRU. We hypothesize that this is due to the need to predict spans. Nevertheless, the $300d$ MRU outperforms an LSTM encoder and remain competitive to a BiLSTM of similar dimensionality. We also observe that LSTM and MRU are complementary. This is made evident by how stacking MRUs over LSTMs can give a performance boost relative to using each encoder separately.

\begin{table*}[t]
  \centering
\small
    \begin{tabular}{|l|cccc|c|}
    \hline
         Model & Bleu-1 & Bleu-4 & Meteor & Rouge-L & Time\\
          \hline
    Seq2Seq$^\dagger$ & 15.89 & 1.26  & 4.08  & 13.15 & N/A\\
    ASR$^\dagger$  \cite{kadlec2016text}   & 23.20  & 6.39  & 7.77  & 22.26 & N/A \\
    BiDAF$^\dagger$ \cite{seo2016bidirectional}& 33.72 & 15.53 & 15.38 & 36.30  &N/A\\
    \hline
    BiAttention ($300d$ LSTM) & 31.18 & 15.34 & 14.42 & 32.95& $\approx$1 hour\\
    BiAttention ($150d$ BiLSTM) &  34.22 & 18.22 & 16.19 & 38.32 & $\approx$1 hour \\
    BiAttention ($300d$ Sim. MRU)  &9.15  &1.69 & 3.95 & 11.16 & 1 min \\
    BiAttention ($300d$ MRU) & 33.28 & 16.15 & 15.84 & 36.65 & 18 mins\\
    BiAttention ($150d$ Hybrid MRU-LSTM) & \textbf{36.55} & \textbf{19.79} & \textbf{17.87} & \textbf{41.44} & $\approx$1 hour\\
    \hline
    \end{tabular}%
  
    \caption{Experimental Results on the NarrativeQA reading comprehension challenge \cite{kovcisky2017narrativeqa} using summaries. $^\dagger$ are baselines reported by \cite{kovcisky2017narrativeqa}.} 
    \label{tab:narrativeqa}%
\end{table*}%

\subsection{Experimental Results on NarrativeQA}
Table \ref{tab:narrativeqa} reports our results on the NarrativeQA benchmark. First, we observe that $300d$ MRU can achieve comparable performance with BiDAF \cite{seo2016bidirectional}. When compared with a BiLSTM of equal output dimensions ($150d$), we find that our MRU model performs competitively, with less than $1\%$ deprovement across all metrics. However, the time cost required is significantly reduced. The performance of our model is significantly better than $300d$ LSTM model while also being significantly faster. Here, we note that Sim. MRU does not produce reasonable results at all, which seems to be in similar vein to results on SearchQA, i.e., a recursive cell that processes word-by-word is mandatory for span prediction. However, our results show that it is not necessary to construct gates in a word-by-word fashion. Finally, the MRU-LSTM significantly outperforms all models, including BiDAF on this dataset. Performance improvement over the vanilla BiLSTM model ranges from $1\%-3\%$ across all metrics, suggesting that MRU encoders are also effective as a complementary neural building block.

\section{Related Work}
A diverse collection of MC datasets such as SQuAD \cite{rajpurkar2016squad} and CNN/DailyMail \cite{hermann2015teaching} are readily available for benchmarking new deep learning models. New datasets have been recently released \cite{kovcisky2017narrativeqa,joshi2017triviaqa,lai2017race,welbl2017constructing}, claiming to involve a greater need for going beyond simple surface-level matching. As such, these datasets often emphasize the extent of compositional and multi-sentence reasoning required to tackle its questions. In the recent years, a wide range of innovation solutions have also been proposed, mainly involving bi-attention \cite{seo2016bidirectional,DBLP:journals/corr/XiongZS16,cui2016attention} and answer pointers \cite{wang2016machine}. Recent work also investigates the notion of multi-hop reasoning \cite{dhingra2016gated,shen2017reasonet,xu2017towards}, reinforcement learning \cite{shen2017reasonet,wang2017r,hu2017mnemonic} and self-matching / self-attention \cite{kundu2018amanda,wang2017gated}. While many of these works use BiLSTMs are standard building blocks, recent work \cite{wei2018fast} attempts a RNN-less model architecture by utilizing components inspired by the Transformer architecture \cite{vaswani2017attention}. Our work is mainly concerned with designing an efficient encoder that is able to capture not only compositional information but also long-range and short-range information. More specifically, our recurrent MRU encoder takes on a similar architecture to Quasi-Recurrent Neural Networks \cite{DBLP:journals/corr/BradburyMXS16} and Simple Recurrent Units \cite{lei2017training}. A recent work, Cross Temporal Recurrent Networks \cite{tay2017cross} extends QRNNs by fusing temporal gates across question-answer pairs. In these models, gates are pre-learned and then applied. However, different from existing models such as QRNNs that convolution layers as gates, we use a block-based contract-and-expand layers for learning gates. Finally, our model also draws inspiration from dilation, in particular dilated RNNs \cite{chang2017dilated} and dilated convolutions \cite{kalchbrenner2016neural}, that intuitively help to model long-range dependencies. 

 \section{Conclusion and Future Work}
We proposed a novel neural architecture, the \textsc{MRU} encoder and an overall bi-attentive model for both MCQ-based and span prediction MC tasks. We apply
it to three MC datasets and achieve competitive performance on all without the use of recurrent layers.
Our proposed method outperforms DFN, an extremely complex model, without using any LSTM or GRU layer. We also remain competitive to AMANDA and BiDAF without any LSTM/GRU. While our proposed encoder
demonstrates promise on reasoning and understanding natural language, we believe that our encoder is generalizable to other domains beyond machine comprehension. However, we defer this prospect to future work. 
\bibliographystyle{plain}
\bibliography{nips2018}

\end{document}